\documentclass{article}
\usepackage{spconf,amsmath,amssymb,booktabs,multirow,graphicx,xcolor}
\usepackage{microtype}
\usepackage{balance}
\usepackage{url}
\usepackage{hyperref}
\hypersetup{hidelinks}
\title{CAUSAL NEURAL SET FILTERING FOR ONLINE MULTI-TARGET TRACKING}

\name{Zhongdi Liu$^{1,*}$ \qquad Huangyu Dai$^{2,*}$}
\address{$^{1}$Hangzhou Applied Acoustics Research Institute, Hangzhou 310023, China\\
$^{2}$Independent Researcher, Hangzhou, China\\
$^{*}$Equal contribution.}

\begin{document}
\ninept
\maketitle

\begin{abstract}

Transformer-based multi-target tracking (MTT) jointly learns data association and state estimation, but MT3/Track-MT3-style trackers repeatedly re-encode measurement windows, incurring redundant computation. We propose Causal Neural Set Filtering (CNSF)\footnote{\href{https://github.com/daihuangyu/CNSF}{Code: https://github.com/daihuangyu/CNSF}}, a neural set filter that encodes only current measurements while carrying past evidence in a structured recursive track state. CNSF combines exclusive Sinkhorn association, association-conditioned Kalman-shaped updates with moment matching, and recurrent Bernoulli lifecycle modeling with measurement-driven birth. These mechanisms impose soft one-to-one constraints, propagate association-induced state uncertainty, and support existence estimation under missed detections and birth--death transitions. On a held-out three-regime simulated test set, CNSF reduces mean GOSPA and T-GOSPA relative to Track-MT3 by 19.3\% and 30.4\%, with 55.9\% fewer parameters and a $3.76\times$ speedup in single-thread CPU inference.

\end{abstract}


\section{Introduction}

Multi-target tracking (MTT) recursively estimates a time-varying target set from measurements affected by clutter, missed detections, uncertain associations, and target birth and death. Classical Bayesian trackers address these uncertainties through explicit recursive state, association, and existence modeling \cite{fortmann1983jpda,reid1979mht,vo2014glmb,garcia2018pmbm}, providing principled probabilistic semantics but potentially incurring substantial complexity as association uncertainty grows.


Neural set trackers offer a different route by learning data association and state estimation jointly. The Multi-Target Tracking Transformer (MT3) and Track-MT3 formulate point-measurement tracking as Transformer-based set prediction with temporal context recovered from measurement windows~\cite{pinto2021mt3,chen2024trackmt3}. Track-MT3 additionally propagates learned query-level information across frames, but overlapping historical measurements are still repeatedly re-encoded. However, its propagated queries do not explicitly carry kinematic uncertainty and target existence within a structured filter state.


To address this gap, we propose Causal Neural Set Filtering (CNSF),
which replaces window-conditioned set prediction with structured
track-state recursion: only current measurements are encoded, while
past evidence is propagated through the recursive track state. CNSF
uses exclusive Sinkhorn association, association-conditioned
Kalman-shaped updates with moment matching, and recurrent Bernoulli
lifecycle modeling with measurement-driven birth. These mechanisms
support the recursion under measurement competition, association
ambiguity, missed detections, and birth--death transitions. On a
held-out three-regime simulated test set, CNSF reduces mean GOSPA
and T-GOSPA by 19.3\% and 30.4\% over Track-MT3, while using 55.9\% fewer parameters and achieving a $3.76\times$ single-thread CPU speedup. It also achieves the lowest mean GOSPA and T-GOSPA among the evaluated methods.

\section{Related Work}

Classical MTT is built on recursive Bayesian estimation. JPDA/MHT address association ambiguity, while RFS-based methods such as $\delta$-GLMB and PMBM model target existence, birth, and death \cite{fortmann1983jpda,reid1979mht,vo2013labeled,vo2014glmb,garcia2018pmbm}. Association marginalization or hypothesis management can become costly as ambiguity grows. Related advances address association and filtering \cite{liu2019deepda,li2022airborne,wei2024bait,cui2025deepaf,ketashvili2025kalman,golan2026aikf,wang2026rmmnet,zhong2025tphd,zhao2026geometric}, including learned Sinkhorn–Kalman coupling \cite{li2024learning}. CNSF integrates soft exclusive association and association-conditioned moment matching with recurrent existence modeling and measurement-driven birth in a structured recursive track state.

A parallel line formulates tracking as learned set prediction
\cite{vaswani2017attention,lee2019settransformer,carion2020detr}.
MT3/Track-MT3 extend this paradigm to point-measurement MTT
\cite{pinto2021mt3,chen2024trackmt3}, with related persistent-query
designs in TrackFormer, MOTR, GTR, TransMOT, and MOTRv2
\cite{meinhardt2022trackformer,zeng2022motr,zhou2022gtr,
ruppel2022transmot,zhang2023motrv2}. Track-MT3 propagates query-level information while re-encoding measurement windows; CNSF instead propagates a structured recursive track state and encodes only current measurements.

\begin{figure*}[t]
\centering
\includegraphics[width=0.98\textwidth]{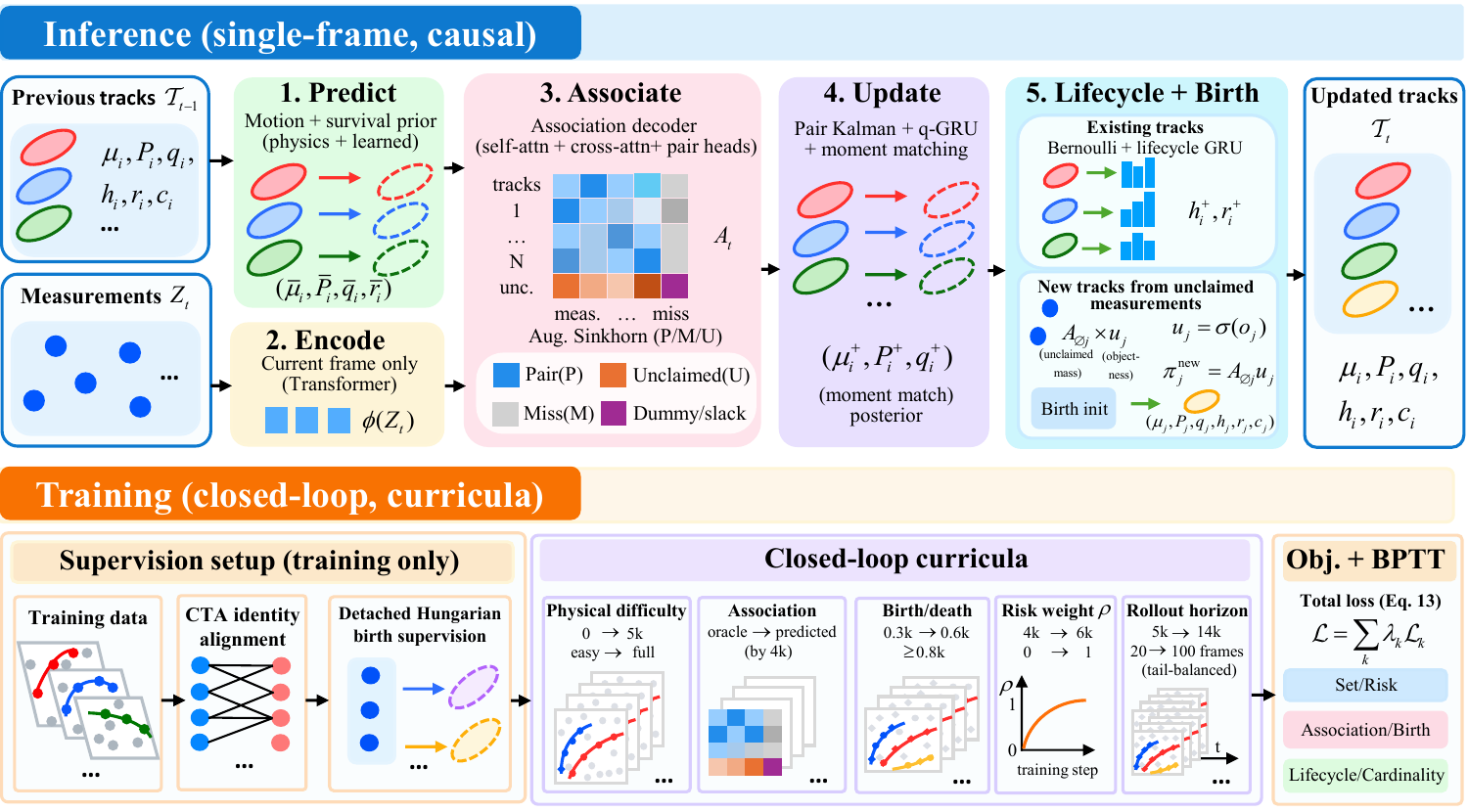}
\caption{Overview of CNSF. \textbf{Top:} causal single-frame inference via prediction, current-frame encoding, exclusive association, moment-matched update, and lifecycle/birth modeling.
\textbf{Bottom:} training-only supervision, closed-loop curricula, and joint BPTT through the recursive path.}
\label{fig:model}
\vspace{-1mm}
\end{figure*}

\section{Causal Neural Set Filter}

The central design choice of CNSF is to make the structured track state, rather than a measurement window, the carrier of temporal information. At each time step, the current measurements $Z_t$ are encoded once and fused with the previous track state $\mathcal T_{t-1}$; the updated state $\mathcal T_t$ is then the only persistent state passed to the next step. This gives the recursive update
\begin{equation}
\mathcal F_\theta:(\mathcal T_{t-1},Z_t)\mapsto\mathcal T_t,
\label{eq:recursion}
\end{equation}
where prediction, association, state update, and target lifecycle
are coupled within a single neural set filter.

\subsection{Recursive state and neural--physical prediction}

Each active track maintains
\begin{equation}
T_i=(\mu_i,P_i,q_i,h_i,r_i,c_i),
\label{eq:track_state}
\end{equation}
where $\mu_i=[x_i,y_i,v_{x,i},v_{y,i}]^\top$ and $P_i$ are the kinematic mean and covariance, $q_i$ is a learned track query, $h_i$ stores lifecycle memory, $r_i\in(0,1)$ is the Bernoulli existence probability, and $c_i$ contains discrete bookkeeping such as age, confirmation, and miss streak. Prediction starts from a constant-velocity model with transition $F(\Delta t)$ and position observation matrix $H=[I_2\;0]$ \cite{kalman1960}. A Fourier embedding $\phi(\Delta t)$ conditions lightweight heads that adapt the query, motion, process uncertainty, and survival probability. With the standard acceleration-input matrix
$G(\Delta t)=[\tfrac12\Delta t^2 I_2\;\;\Delta t I_2]^\top$, prediction is
\begin{equation}
\begin{aligned}
\bar q_i   &= q_i+\Delta q_i,
&\qquad \bar\mu_i &= F\mu_i+G a_i,\\
\bar P_i   &= FP_iF^\top+Q_i,
&\qquad \bar r_i  &= r_i\,\sigma\!\bigl(\operatorname{logit}(p_0^S)+\Delta s_i\bigr).
\end{aligned}
\label{eq:prediction}
\end{equation}
Here $\Delta q_i=f_\theta^q(q_i,\phi(\Delta t))$ and
$a_i=f_\theta^a(\bar q_i)$. The learned process covariance is
$Q_i=\operatorname{Diag}(\operatorname{softplus}(f_\theta^Q(\bar q_i))+\epsilon)\Delta t$,
and the survival correction is bounded as
$\Delta s_i=\kappa_S\tanh(f_\theta^S/\kappa_S)$, where $f_\theta^S$ denotes the pre-measurement survival head conditioned only on predicted track and causal lifecycle features. In parallel, a permutation-equivariant Transformer encodes the current measurement set $Z_t=\{z_t^j\}_{j=1}^{M_t}$ once; historical measurements are not re-encoded.

\subsection{Exclusive association and uncertainty propagation}

For track--measurement pair $(i,j)$, define the innovation
$\nu_{ij}=z_t^j-H\bar\mu_i$ and its covariance
$S_{ij}=H\bar P_iH^\top+R_{ij}$, where the pair head predicts a
positive diagonal observation covariance $R_{ij}$. Let
$\psi_{ij}=f_\theta^{\rm pair}(i,j)+b_i^\theta$ denote learned pair
evidence and $g_{ij}=\sigma(f_\theta^{g}(i,j))\in(0,1)$ a learned
gate on the physical compatibility score:
\begin{equation}
\begin{aligned}
\ell^{\rm phy}_{ij}
&=-\frac12\!\left(
\nu_{ij}^\top S_{ij}^{-1}\nu_{ij}+\log|S_{ij}|
\right),\\[-1mm]
s_{ij}
&=g_{ij}\ell^{\rm phy}_{ij}+\psi_{ij}
+\lambda_{\rm prior}\operatorname{logit}\!\bigl(
\max(\bar r_i,r_{\min})
\bigr).
\end{aligned}
\label{eq:association_score}
\end{equation}

Let $N$ and $M$ be the numbers of active tracks and valid
measurements. We augment the pair scores with learned MISS logits
$s_{i\varnothing}$, UNCLAIMED logits $s_{\varnothing j}$, and a
slack entry, forming $S^{\rm aug}\in\mathbb R^{(N+1)\times(M+1)}$.
Association masses are obtained by
\[
A=\operatorname{Sinkhorn}
\!\left(S^{\rm aug}/\tau_{\rm sk};\,\mathbf a,\mathbf b\right),
\]
using $N_{\rm sk}$ log-space iterations \cite{cuturi2013sinkhorn},
with marginals
$\mathbf a=[\mathbf 1_N;M]$ and
$\mathbf b=[\mathbf 1_M;N]$.
Thus, for active tracks and valid measurements,
$\sum_{j=1}^{M}A_{ij}+A_{i\varnothing}=1$ and
$\sum_{i=1}^{N}A_{ij}+A_{\varnothing j}=1$,
enforcing soft one-to-one association constraints; the corner entry serves as the slack needed by the augmented marginals.

Each PAIR hypothesis is updated before marginalization. The same
learned $R_{ij}$ used for association enters a Kalman-shaped update
with a bounded gain residual:
\begin{align}
K_{ij}
&=\bar P_iH^\top S_{ij}^{-1}
+\kappa_K\tanh(\Delta K_{ij}^\theta),
\qquad
\mu_{ij}=\bar\mu_i+K_{ij}\nu_{ij},
\nonumber\\
P_{ij}
&=(I_4-K_{ij}H)\bar P_i(I_4-K_{ij}H)^\top
+K_{ij}R_{ij}K_{ij}^\top .
\label{eq:pair_update}
\end{align}
The pair representation also updates the latent query through a gated recurrent unit (GRU), yielding $q_{ij}$. For each active track, we renormalize its Sinkhorn row to obtain
$w_{ij}$ and $w_{i0}$, with
$(\mu_{i0},P_{i0},q_{i0})=(\bar\mu_i,\bar P_i,\bar q_i)$.
The pair and MISS hypotheses are then collapsed into one recursive
state by moment matching:
\begin{align}
\mu_i^+
&=\sum_{k=0}^{M}w_{ik}\mu_{ik},
\qquad
q_i^+=\sum_{k=0}^{M}w_{ik}q_{ik},
\nonumber\\
P_i^+
&=\sum_{k=0}^{M}w_{ik}
\!\left[
P_{ik}
+(\mu_{ik}-\mu_i^+)(\mu_{ik}-\mu_i^+)^\top
\right].
\label{eq:moment}
\end{align}
Eq.~\eqref{eq:moment} preserves the mixture mean and covariance, including the between-component covariance induced by association ambiguity, while retaining a single moment-matched state per track rather than propagating multiple association hypotheses.

\subsection{Bernoulli lifecycle and network architecture}

Let $u_j=\sigma(o_j)$ denote measurement objectness and
$m_i=\sum_j w_{ij}u_j$ the objectness-weighted PAIR support for track $i$.
Lifecycle memory is updated from the updated query and causal evidence,
while a bounded learned correction acts in existence log-odds space:
\begin{equation}
\begin{array}{@{}l@{\;}c@{\;}l@{}}
h_i^+ & = & \operatorname{GRU}\!\left([q_i^+,\xi_i],h_i\right),\\[1pt]
r_i^+ & = & \sigma\!\left(
\operatorname{logit}\bar r_i+m_i-\delta_m+\Delta\ell_i^r
\right),\\[1pt]
\pi_j^{\rm new} & = & A_{\varnothing j}u_j ,
\end{array}
\label{eq:lifecycle}
\end{equation}
where
$\Delta\ell_i^r=\kappa_r\tanh(f_\theta^r(h_i^+,\xi_i')/\kappa_r)$.
The feature vector $\xi_i$ contains previous and instantaneous existence, PAIR/MISS support, objectness support, association entropy and margin, elapsed time, updated covariance, and age; $\xi_i'$ retains the PAIR support, ambiguity, time, and covariance terms used by the log-odds correction head. Birth heads initialize position, velocity, diagonal covariance, query, lifecycle memory, and existence. The updated existence $r_i^+$ then controls confirmation, retention, and termination.

Measurement/track-query features have width $d$ and lifecycle memory width $d_h$. The measurement encoder uses $L_e$ layers, $H$ attention heads, and feed-forward width $d_{\rm ff}$; the $L_d$-layer association decoder applies track self-attention followed by track-to-measurement cross-attention. Lightweight heads parameterize Eqs.~\eqref{eq:prediction}--\eqref{eq:lifecycle}.

\subsection{Training objective}
Training preserves recursive identities across frames. Existing tracks inherit their ground-truth identities through cross-frame target alignment (CTA), while Hungarian assignment is used only to match unmatched birth candidates to unmatched targets during training. The discrete assignment is detached and never used at inference. Losses are computed per supervised frame and averaged across valid rollout frames. Here $\operatorname{BCE}(\ell,y)$ denotes binary cross-entropy with logits.

\textbf{Set loss.} Let $\mathcal M_t$ be matched prediction--target pairs and $\mathcal C_t$ all valid existing/birth candidates. Candidate $k$ predicts position $\hat p_k$, logit $\ell_k$, and label $y_k\in\{0,1\}$; $p_n^*$ is the matched target position:
\begin{equation}
\mathcal L_{\rm set}=\frac{1}{|\mathcal M_t|}\!\sum_{(k,n)\in\mathcal M_t}\frac12\|\hat p_k-p_n^*\|_1
+\frac{1}{|\mathcal C_t|}\!\sum_{k\in\mathcal C_t}\operatorname{BCE}(\ell_k,y_k).
\label{eq:setloss}
\end{equation}
An empty matched set contributes zero localization loss.

\textbf{Probabilistic set-risk surrogate.} Reusing the detached CTA/Hungarian matching, define $\pi_k=\sigma(\ell_k)$, cutoff $c_{\rm risk}$, $d_c(k,n)=\min(\|\hat p_k-p_n^*\|_2,c_{\rm risk})$, and unmatched prediction/truth sets $\mathcal U_b,\mathcal V_b$:
\begin{align}
\mathcal L_{\rm risk}=\frac1B\sum_b\Bigg[&\sum_{(k,n)\in\mathcal M_b}
\left(\pi_k d_c(k,n)+(1-\pi_k)\frac {c_{\rm risk}}2\right)\nonumber\\
&+\sum_{k\in\mathcal U_b}\pi_k\frac {c_{\rm risk}}2+|\mathcal V_b|\frac {c_{\rm risk}}2\Bigg].
\label{eq:risk}
\end{align}
The matching is treated as fixed supervision rather than recomputed to minimize this risk. Its coefficient $\rho$ follows the curriculum in Sec.~\ref{sec:experiments}.

\textbf{Association and birth losses.} Simulator identities define the correct PAIR, MISS, and UNCLAIMED event sets $E_p,E_m,E_u$. With event weights $w_p,w_m,w_u$ and $Z_E$ equal to the sum of weights of event types present in the frame,
\begin{align}
\mathcal L_{\rm assoc}=-\frac1{Z_E}\Bigg[&\frac{w_p}{|E_p|}\sum_{(i,j)\in E_p}\log A_{ij}
+\frac{w_m}{|E_m|}\sum_{i\in E_m}\log A_{i\varnothing}\nonumber\\
&+\frac{w_u}{|E_u|}\sum_{j\in E_u}\log A_{\varnothing j}\Bigg].
\label{eq:assocloss}
\end{align}
Here, absent event types are omitted together with their weight. For birth, when both classes are present,
$\mathcal B(p,y)=-\tfrac12[\operatorname{mean}_{y=1}\log p+\operatorname{mean}_{y=0}\log(1-p)]$;
if only one class is present, $\mathcal B$ is the corresponding class mean. Then
\begin{equation}
\mathcal L_{\rm birth}=\mathcal B(\{u_j\},\{y_j^{\rm tgt}\})+\mathcal B(\{\pi_j^{\rm new}\},\{y_j^{\rm new}\}).
\label{eq:birthloss}
\end{equation}

\textbf{Lifecycle and cardinality.} For active slots $\mathcal A_t$, $y_i^{\rm alive}$ indicates current truth existence and $y_i^{\rm det}$ current detection; live-but-missed tracks use $\omega_i=1+\lambda_{\rm miss}\mathbf1[y_i^{\rm alive}=1,y_i^{\rm det}=0]$. For $\mathcal S_t\subseteq\mathcal A_t$ that existed at $t-1$, $y_i^S$ is one-step survival. With $p_i^S=\sigma(\operatorname{logit}(p_0^S)+\Delta s_i)$, $\ell_i^{\rm exist}=\operatorname{logit}r_i^+$, $\ell_i^S=\operatorname{logit}p_i^S$, and $W_t=\sum_{i\in\mathcal A_t}\omega_i$,
\begin{equation}
\begin{array}{@{}l@{\;}c@{\;}l@{}}
\mathcal L_{\rm exist} & = & \displaystyle \frac1{W_t}\sum_{i\in\mathcal A_t}
\omega_i\operatorname{BCE}(\ell_i^{\rm exist},y_i^{\rm alive}),\\[2pt]
\mathcal L_{\rm surv} & = & \displaystyle \frac1{|\mathcal S_t|}\sum_{i\in\mathcal S_t}
\operatorname{BCE}(\ell_i^S,y_i^S),\\[2pt]
\hat N_t & = & \displaystyle \sum_i m_i^{\rm ex}r_i^+ +\sum_jm_j^{\rm b}\pi_j^{\rm new},\\[2pt]
\mathcal L_{\rm card} & = & \operatorname{SmoothL1}(\hat N_t,N_t^*).
\end{array}
\label{eq:lifecycleloss}
\end{equation}
where $m^{\rm ex},m^{\rm b}$ are validity masks and $N_t^*$ is the true cardinality. Empty active/survival sets skip the corresponding term. The complete objective is
\begin{align}
\mathcal L={}&\lambda_{\rm set}\mathcal L_{\rm set}+\lambda_{\rm risk}\rho\mathcal L_{\rm risk}
+\lambda_{\rm assoc}\mathcal L_{\rm assoc}+\lambda_{\rm birth}\mathcal L_{\rm birth}\nonumber\\
&+\lambda_{\rm exist}\mathcal L_{\rm exist}+\lambda_{\rm surv}\mathcal L_{\rm surv}
+\lambda_{\rm card}\mathcal L_{\rm card}.
\label{eq:total}
\end{align}
We use $c_{\rm risk}=2$, $(w_p,w_m,w_u)=(2,1,1)$, and
$\lambda_{\rm miss}=2$. The loss weights
$(\lambda_{\rm set},\lambda_{\rm risk},\lambda_{\rm assoc},
\lambda_{\rm birth},\lambda_{\rm exist},\lambda_{\rm surv},
\lambda_{\rm card})$ are $(1,.1,1,.5,.35,.25,.03)$,
with $\rho$ ramped over steps 4k--6k.

\section{Experiments}
\label{sec:experiments}
\subsection{Protocol and training configuration}

\begin{figure*}[!t]
\vspace{-3mm}
\centering
\includegraphics[width=0.98\textwidth]{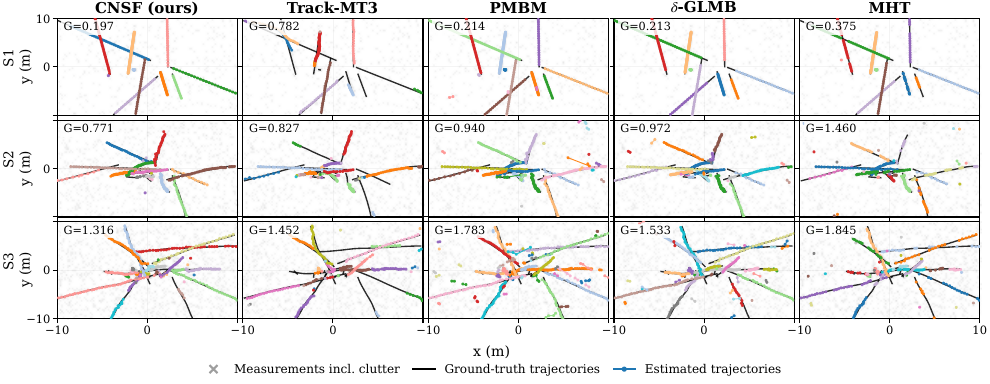}
\caption{Fixed qualitative cases (run 000): three scenarios (rows) $\times$ five methods (columns), with identical measurements and ground truth. $G$ is mean GOSPA over frames 20--99; axes are in meters.}
\label{fig:tracking}
\vspace{-4mm}
\end{figure*}

We evaluate three 100-frame simulated point-target regimes of increasing difficulty with $\Delta t=0.1$ over $[-10,10]^2$. Initial and newborn positions and velocities are drawn independently from $\mathcal N(0,3I_2)$; targets are removed upon survival failure or field-of-view exit. For S1--S3, $(\lambda_B,p_S,p_D,q,r,\lambda_C)$ are $(.04,.99,.95,0,0,5)$, $(.08,.98,.90,.04,.02,10)$, and $(.12,.97,.85,.08,.03,15)$, respectively, denoting birth rate, survival probability, detection probability, process-noise intensity, measurement-noise intensity, and clutter rate. Here $q$ scales the CV white-acceleration covariance, and the measurement covariance is $rI_2$. The regimes start with 6/6/10 targets, cap truth cardinality at 16, and use uniform clutter. The held-out test set contains 50 trajectories per regime; frames 20--99 are scored using GOSPA $(p=1,c=2,\alpha=2)$ \cite{rahmathullah2017gospa}, Probabilistic GOSPA (Pro-GOSPA) \cite{xia2025pgospa}, and T-GOSPA $(p=1,c=2,\gamma=1)$ \cite{garcia2020tgospa}. T-GOSPA is normalized over the 80 scored frames and uses tracker-native identities without post-hoc relinking; Track-MT3 variants use persistent cross-frame query identities.

\begin{table}[t]
\caption{Held-out test performance. $G/P/T$: mean
GOSPA/Pro-GOSPA/T-GOSPA; CPU: ms/frame; Par.: millions.
$\dagger$ For deterministic point outputs with unit existence, Pro-GOSPA equals GOSPA.}
\label{tab:main}
\vspace{1mm}
\centering
\renewcommand{\arraystretch}{1.10}
\setlength{\tabcolsep}{1.0pt}
\begin{tabular}{@{}lrrrrrrrr@{}}
\toprule
Method & S1 & S2 & S3 & $G$ & $P$ & $T$ & CPU & Par. \\
\midrule
MHT & .576 & 1.601 & 2.579 & 1.585 & 1.585$^{\dagger}$ & 1.676 & 89.04 & -- \\
$\delta$-GLMB & .250 & .998 & 2.034 & 1.094 & 1.094$^{\dagger}$ & 1.251 & 1614.57 & -- \\
PMBM & .270 & .982 & 1.945 & 1.066 & 1.250 & 1.179 & 81.47 & -- \\
Track-MT3 & .609 & .990 & 1.680 & 1.093 & 1.245 & 1.443 & 117.37 & 19.38 \\
Track-MT3-CM & .642 & 1.050 & 1.750 & 1.148 & 1.427 & 1.595 & 39.68 & 8.48 \\
\textbf{CNSF} & \textbf{.166} & \textbf{.828} & \textbf{1.652} &
\textbf{.882} & \textbf{.990} & \textbf{1.004} & \textbf{31.18} & 8.54 \\
\bottomrule
\end{tabular}
\end{table}

A separate 10-trajectory validation set is used only for checkpoint and global operating-point selection. Learned trackers save checkpoints every 2k updates from step 8k; the checkpoint--operating-point pair with the lowest validation mean GOSPA is frozen for test evaluation, without per-scene or per-frame tuning. Baselines include MHT, $\delta$-GLMB, PMBM, and Track-MT3 with a 20-frame window and six-layer encoder/decoder. A capacity-matched variant, Track-MT3-CM, retains the same 20-frame formulation while matching CNSF in model capacity (8.48M vs.\ 8.54M) with its 12k checkpoint selected by the same validation protocol.

Classical filters use the true regime-specific motion/noise, detection, survival (where applicable), and clutter parameters, with measurement-driven birth. MHT/$\delta$-GLMB use fixed 8/64 hypothesis caps, while PMBM uses 20/25 desired/maximum hypotheses; these settings are fixed across regimes.

CNSF uses $d=d_h=256$, $L_e=L_d=3$, $H=8$, $d_{\rm ff}=1024$, 32 active slots, $N_{\rm sk}=20$, and $\tau_{\rm sk}=0.75$, totaling 8.54M parameters. Validation-selected birth-candidate/immediate-birth/existing-output/retention thresholds are $0.325/0.68/0.55/0.225$ and frozen for testing. Training generates trajectories on the fly across the three regimes and uses AdamW for 20k updates with batch size 48 and peak learning rate $1.2\times10^{-4}$; physical difficulty reaches its full regime by step 5k, association becomes fully predicted by step 4k, and the backpropagation-through-time (BPTT) horizon grows from 20 to 100 frames by step 14k. Inference remains one-frame recursive. CPU latency uses one thread, batch size one, the same S3 sequence, five repeats, and a common input-to-output timing boundary.

\subsection{Results and controlled ablations}



Table~\ref{tab:main} summarizes the main results. CNSF achieves the lowest mean GOSPA and T-GOSPA among the evaluated methods. Relative to Track-MT3, the two metrics decrease by 19.3\% and 30.4\%, while measured single-thread CPU latency decreases from 117.37 to 31.18 ms/frame with 55.9\% fewer parameters. The capacity-matched Track-MT3-CM (8.48M) yields G/T = 1.148/1.595, versus .882/1.004 for CNSF (8.54M). CNSF also reduces latency from 39.68 to 31.18 ms/frame (21.4\%; $1.27\times$ speedup), so the observed advantage persists at comparable parameter counts.

The gain is strongest in S1/S2, while CNSF remains competitive in the dense S3 regime. There, CNSF and Track-MT3 have similar frame-wise GOSPA (1.6518 vs.\ 1.6803). CNSF achieves lower localization error (0.6471 vs.\ 0.7378), but slightly higher missed/false penalties (0.7585/0.2463 vs.\ 0.7313/0.2113). Thus, its S3 GOSPA gain comes mainly from improved localization rather than cardinality handling. CNSF also achieves lower S3 T-GOSPA (1.8837 vs.\ 2.218), so the advantage extends to trajectory-level evaluation. Figure~\ref{fig:tracking} shows the same fixed sequence for all methods in each regime.

\begin{table}[t]
\caption{Structural ablations on the held-out test set. $\bar G/\bar P/\bar T$: mean GOSPA/Pro-GOSPA/T-GOSPA; $M_3/F_3$: S3 missed/false.}
\label{tab:ablation}
\vspace{1mm}
\centering
\renewcommand{\arraystretch}{1.13}
\setlength{\tabcolsep}{3.5pt}
\begin{tabular}{@{}lccccc@{}}
\toprule
Variant & $\bar G$ & $\bar P$ & $\bar T$ & $M_3$ & $F_3$ \\
\midrule
Full CNSF & \textbf{.882} & \textbf{.990} & \textbf{1.004} & \textbf{.759} & .246 \\
w/o excl. Sinkhorn & 1.524 & 1.539 & 1.659 & .724 & 1.325 \\
w/o pairwise mix. & .931 & 1.053 & 1.060 & .847 & .250 \\
w/o rec. lifecycle & 1.112 & 1.194 & 1.248 & 1.195 & \textbf{.177} \\
\bottomrule
\end{tabular}
\end{table}

Table~\ref{tab:ablation} compares three structural variants. \emph{w/o excl. Sinkhorn} uses independent row softmax; \emph{w/o pairwise mix.} replaces pairwise state mixing with a single expected-measurement update; and \emph{w/o rec. lifecycle} uses an instantaneous existence head. The Sinkhorn replacement yields the largest degradation ($\bar T$: 1.004$\rightarrow$1.659), increasing S3 false error from 0.246 to 1.325. The lifecycle replacement raises $\bar T$ to 1.248 and S3 missed error to 1.195, but lowers false error to 0.177. Replacing pairwise mixing increases all three mean metrics, with a smaller rise in $\bar T$ (1.004$\rightarrow$1.060).

\section{Conclusion}
We presented CNSF, an online neural set filter that encodes only current measurements and propagates history through a structured recursive track state. On the held-out test set, it achieves the lowest mean GOSPA/T-GOSPA among the evaluated methods and outperforms capacity-matched Track-MT3-CM in accuracy and speed. Its main limitation is compressing prolonged association ambiguity into a single moment-matched state.

\clearpage
\balance

\bibliographystyle{IEEEbib}
\bibliography{references}

\end{document}